\documentclass{article}

\PassOptionsToPackage{numbers, compress}{natbib}

    \usepackage[preprint]{neurips_2020}

\usepackage[utf8]{inputenc} 
\usepackage[T1]{fontenc}    
\usepackage{hyperref}       
\usepackage{url}            
\usepackage{booktabs}       
\usepackage{amsfonts}       
\usepackage{nicefrac}       
\usepackage{microtype}      
\usepackage{algorithm}
\usepackage{algorithmic}
\usepackage{forloop}
\usepackage{xcolor}

\usepackage{graphicx}

\usepackage{color}
\usepackage{multirow}
\usepackage{array}
\usepackage{wrapfig}

\title{Disentangle Perceptual Learning through Online Contrastive Learning}

\author{%
  Kangfu Mei$^{1}$,~~Yao Lu$^{1}$,~~Qiaosi Yi$^{1}$,~~Haoyu Wu$^{1}$, ~~Juncheng Li$^{2}$, ~~Rui Huang$^{1}$\\
    $^{1}$The Chinese University of Hong Kong, Shenzhen \\ $^{2}$East China Normal University\\
  \texttt{kangfumei@link.cuhk.edu.cn}
}

\begin{document}

\maketitle

\begin{abstract}
Pursuing realistic results according to human visual perception is the central concern in the image transformation tasks.
Perceptual learning approaches like~\cite{johnson2016perceptual} are empirically powerful for such tasks but they usually rely on the pre-trained classification network to provide features, which are not necessarily optimal in terms of visual perception of image transformation.
In this paper, we argue that, among the features representation from the pre-trained classification network, only limited dimensions are related to human visual perception, while others are irrelevant, although both will affect the final image transformation results.
Under such an assumption, we try to disentangle the perception-relevant dimensions from the representation through our proposed online contrastive learning.
The resulted network includes the pre-training part and a feature selection layer, followed by the contrastive learning module, which utilizes the transformed results, target images, and task-oriented distorted images as the positive, negative, and anchor samples, respectively.
The contrastive learning aims at activating the perception-relevant dimensions and suppressing the irrelevant ones by using the triplet loss, so that the original representation can be disentangled for better perceptual quality.
Experiments on various image transformation tasks demonstrate the superiority of our framework, in terms of human visual perception, to the existing approaches using pre-trained networks and empirically designed losses.
\end{abstract}

\section{Introduction}
Image transformation aims at transforming images from one condition/scenario into another, e.g., low-resolution into high-resolution, low-lighting into normal-lighting, etc.
Recent deep learning-based methods~\cite{dong2015image}\cite{lim2017enhanced}\cite{zhang2018image} have achieved significant improvements in transforming the contents of images, but the visual quality of the transformed images are often not perfect, especially in terms of human perception.

More recent works~\cite{ledig2017photo}\cite{wang2018esrgan}\cite{jolicoeur2018relativistic} introduce perceptual learning to address this issue.
They utilize a pre-trained classification network \(\Psi\) to extract high-dimension features as the representations of both the generated images \(\tilde{X}\) and the target images \(Y\), and then measure the distance between these two representations as the loss function, formulated as:
\begin{equation}
  \mathcal{L}_{perceptual }(\tilde{X}, Y) = || \Psi(\tilde{X}) - \Psi(Y)||^2.
\end{equation}
Compared with the distance metrics like MAE or MSE used in earlier works, the perceptual distance is measured in the feature space instead of the pixel space, which is considered to be more compact and more relevant to human perception. 
Furthermore, Mechrez et al.~\cite{mechrez2018contextual} propose the contextual loss, which measures the distance in a feature contextual space, defined as:
\begin{equation}
\mathcal{L}_{CX}(\tilde{X}, Y) = -\log{(\frac{1}{N} \sum_j \max_i \mbox{CX}_{ij})},
\end{equation}
where \(\mbox{CX}_{ij}\) is the similarity between features \(\Psi(\tilde{X})={\tilde{x}_i}\) and \(\Psi(Y)={y_j}\), and is usually calculated using the normalized cosine distance.
Compared to \(\mathcal{L}_{perceptual}(\tilde{X}, Y)\), \(\mathcal{L}_{CX}(\tilde{X}, Y)\) is calculated in the feature contextual space, which is supposed to be more robust when the training images \(\tilde{X}\) and \(Y\) are not aligned.
The impact of these methods is analyzed in more detail in Mechrez et al.~\cite{mechrez2018maintaining} and Yang et al.~\cite{yang2019deep}.

In general, the extracted features from pre-trained networks can be regarded as probability distributions over the input images \( \tilde{X} \) and \( Y \), denoted as \( P_{\tilde{X}} \) and \( P_Y \).
Then we can demonstrate that minimizing the distance between \(\Psi(\tilde{X})\) and \(\Psi(Y)\) using \(\mathcal{L}_{perceptual}\) or \(\mathcal{L}_{CX}\) is similar to minimizing the Kullback-Leibler (KL) divergence between two distributions \(P_{\tilde{X}}\) and \(P_Y\), as shown in Eq.(3):
\begin{equation}
    D_{KL} (P_{\tilde{X}}||P_Y) = \int{P_{\tilde{X}} \log \frac{P_{\tilde{X}}}{P_Y}.}
\end{equation}

Since \(\Psi(\tilde{X})\) and \(\Psi(Y)\) are extracted features from the pre-trained network \(\Psi\) (usually a classification network trained on large-scale classification datasets), we can consider \(P_{\tilde{X}}\) and \(P_Y\) as the mappings from the pixel space into the semantic manifold of nature images learned by the pre-trained network \(\Psi\).
Therefore, the generated images using these perceptual learning approaches can be more realistic.

\begin{figure*}[t]
    \centering
    \includegraphics[scale=0.15]{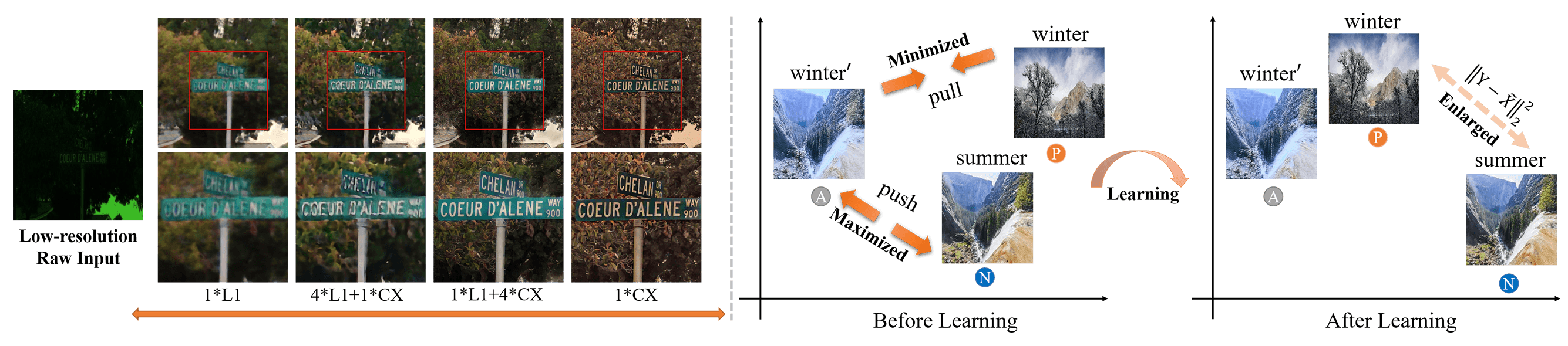}
    \vspace{-14px}
    \caption{The left part: the image quality tradeoff between the contextual loss and L1 loss. The right part: changes in feature distance between different seasons before and after learning.}
    \label{fig:tradeoff}
    \vspace{-14px}
\end{figure*}

It should be noted, however, these two loss functions often account for relatively small roles in the final loss function of the entire network training, even though they might be important for human perception.
In practice, they are often combined with the traditional pixel-wise losses or adversarial losses, and only worked as the auxiliary to avoid artifacts.
We argue that these perceptual losses do not work well alone because of the irrelevant features contained in the pre-trained representations and the poor generalization ability of the feature projections when used on new datasets.
For the first problem, we conduct several experiments in combining the contextual loss with L1 loss with different weights, and observe that directly using the contextual loss will lead to unexpected artifacts in the generated images, e.g., color offset, ripple artifacts, blurring, etc.
These results are shown in the left part of Figure~\ref{fig:tradeoff}.
Increasing the weight of the L1 loss will reduce the artifacts but cause blur in the generated images.
Furthermore, in the middle part of figure~\ref{fig:tradeoff}, we visualize the dimension-reduced representation of three images from two scenes and two seasons mapped using the pre-trained networks.
We notice that these two images from the same scene but different seasons have a smaller distance than that of the two images from different scenes but the same season, which means the pre-trained features are not suitable to the transformation tasks like season transfer.
To achieve season transfer, we need to push the distance between the images of different seasons away and pull the distance between the same-season images closer, as shown in the right part of figure~\ref{fig:tradeoff}. 

In other words, we believe that features pre-trained for classification tasks contain information that is irrelevant to other image transformation tasks.  The perceptual losses designed upon these features, though capture additional perceptual information than the simple pixel-wise losses, also introduce many irrelevant information that misleads the training of the image transformation networks.
Such features, therefore, should not be directly applied in the image transformation network training process.
To overcome this drawback, in this paper we propose to fine-tune the pre-trained network using the task-oriented instance triplet samples, so that the traditional pre-trained representations can be disentangled as a set of task-relevant dimensions and a set of task-irrelevant dimensions.
To achieve this goal, we introduce a novel online contrastive learning scheme to activate the task-relevant dimensions and suppress the task-irrelevant dimensions.

\section{Related Work}
\subsection{Image Transformation}
Pursuing realistic image transformation has been addressed by several recent works.
Both adversarial based methods, e.g., SRGAN~\cite{ledig2017photo} and perceptual based methods, e.g., Perceptual Loss~\cite{johnson2016perceptual} are proposed to push the generated images' realism improvements by minimizing the distance between high-dimension features, which represent either semantic or classic information from the pre-training or parallel-training deep networks.
Compared with the pixel-based loss function, e.g., MAE and MSE Losses, usually leading to over-smooth results, these works tend to generate finer texture details.
However, they suffer ambiguous convergence during adversarial training or unpleasant artifacts, mainly caused by the unreasonable deep features represent both realistic relevant dimensions and non-relevant dimensions.
To better reduce the artifacts, some works, e.g., ESRGAN~\cite{wang2018esrgan} have resorted to select stronger features from the pre-training network to represent the images, and enhance the network architecture on generation and discrimination to reduce the convergence difficulties.
Different from the representation enhancement, the idea of optimizing the distance measuring has also been discussed in recent works, e.g., Contextual Loss~\cite{mechrez2018contextual} computes the similarities on the contextual features, which could be seen as maximizing the similarities with images' real distributions via non-parametric estimation, instead of minimizing the Euclidean distance as discussed previously.
Even these methods are robustness on unaligned data due to the optimized distance measuring and tend to generate pleasant results, there still existed many failed cases of the results.
Moreover, these methods lack of convincing reasons for the feature selection or effectiveness analysis of the realistic, which further increase the difficulties of improvement in themselves and can only be applied as a black-box.

\subsection{Representation Learning}
Extracting efficient representation from images is crucial to measuring the distance between transformed images and target images.
Although Perceptual Loss~\cite{johnson2016perceptual} and Contextual Loss~\cite{mechrez2018contextual} have declared that the features extracted from the pre-trained VGG-19 or AlexNet can best represent the human perception in images, there still have many works proposed to refine representations from pre-training.
Using networks pre-trained in unsupervised common tasks and then fine-tune representations in specific sub-tasks have shown its superiority in both vision and language tasks, e.g., MoCo~\cite{he2019momentum}, SimCLR~\cite{chen2020simple}, GPT~\cite{radford2018improving}, and BERT~\cite{devlin2018bert}.
These methods use contrastive learning to pre-train on common tasks so that they can capture the relations between similar objects and dissimilar relations without labels.
For example, triplet loss~\cite{hoffer2015deep}\cite{schroff2015facenet} maximizes the distance from anchor samples to negative samples and minimizes the distance from anchor samples to positive samples.
Under such a process, the optimized networks can learn useful representations from data.
However, these vision-related pre-training methods are usually applied to the recognition~\cite{chen2020simple} related tasks rather than generation related tasks.
Regarding to the image generation related tasks, most works use the representation of images as the latent code between the encoder and decoder.
Then discriminator is used to regularize the generated results with specific latent code.
Chen et al.~\cite{chen2016infogan} introduce InfoGAN that uses unsupervised learning to learn disentangled representations, which decomposes the input into an incompressible noise and a latent code so that the representation is related to the latent code only.
Even these methods can generate diversified results using the encoded representation, the quality of generated results cannot be ensured.
Jolicoeur et al.~\cite{jolicoeur2018relativistic} and Wang et al.~\cite{wang2018esrgan} apply the probabilities of generated images to stable the training process and enhance the quality of images.
The probability similarities are measured in the feature differences of the discriminator, which could be seen as the special case of measuring representation differences.

\section{Disentangled Perceptual Learning}
\begin{algorithm*}[t]
\caption{\label{alg:main} our proposed Disentangled Perceptual Learning.} 
\begin{algorithmic}
    \STATE \textbf{input:} source images $X$, target images $Y$, generator network $F(\cdot)$, pre-trained network $\Psi(\cdot)$, feature selection layer $\Phi(\cdot)$, accumulate interval $N$, random crop function $f_c$, task-oriented distortion function $f_d$.
    \FOR{sampled mini-batch $\{x_k\}_{k=1}^N$, $\{y_k\}_{k=1}^N$}
    \STATE \textbf{for all} $k\in \{1, \ldots, N\}$ \textbf{do}
        \STATE $~~~~$ \textbf{do} freeze parameters of $F$
        \STATE $~~~~$ $\tilde{x}_k = F(x_k)$
        \STATE $~~~~$ $<y', y'', \tilde{x}'>_k = f_c(y_k), f_c(f_d(y_k)), f_c(\tilde{x}_k)$
        \STATE $~~~~$ $<h_n, h_a, h_p>_k = \Psi (<y', y'', \tilde{x}'>_k)$
        \STATE $~~~~$ $<e_n, e_a, e_p>_k = \Phi (<h_n, h_a, h_p>_k)$ 
        \STATE $~~~~$\textcolor{gray}{\# accumulate gradient of loss $d_c$}
        \STATE $~~~~$ $d_c \leftarrow d_c + \max( || e_a - e_p||^2_2 - || e_a - e_n ||^2_2 + \mbox{margin}, 0)$
        
        \STATE $~~~~$ $<h_n, h_p>_k = \Psi(<y, \tilde{x}>_k)$
        \STATE $~~~~$ $<e_n, e_p>_k = \Phi(<h_n, h_p>_k)$
        \STATE $~~~~$ \textbf{do} unfreeze parameters of $F$
        \STATE $~~~~$ $F \leftarrow	F + \mbox{Adam}(F, ||e_n - e_p||^2)$
        \STATE \textbf{end for}
        \STATE $\Phi \leftarrow	\Phi + \mbox{Adam}(\Phi, d_c)$
    \ENDFOR
    \STATE \textbf{return} generator network $F(\cdot)$, and throw away $\Psi(\cdot)$ and $\Phi(\cdot)$
\end{algorithmic}
\end{algorithm*}

In this paper, we adapt the perceptual loss as the main loss function in training networks, which uses the deep feature-based loss instead of pixel loss, adversarial loss, or any other handcrafted loss functions.
Deep feature-based loss function tends to generate images with more realistic details and provides a more stable training process empirically, but it is usually incorporated
as an auxiliary loss function in previous works
due to its inexplicable and uncontrollable problems.
To overcome these issues, in this section we describe the details of our proposed Disentangled Perceptual Learning (DPL) with online contrastive learning as a new general framework of perceptual learning.
We separate the DPL into three different components: Online Contrastive Learning, Feature Selection as Fine-tune, and Task-Oriented Disentanglement.
We summarize the overall method in Algorithm~\ref{alg:main}.

\subsection{Online Contrastive Learning}
The superiority of perceptual learning mainly comes from the applied pre-trained classification network.
A classification network \( \Psi \) pre-trained on a large-scale image classification dataset can map the input image into a high-level feature space, where images with similar contents will be projected into similar embeddings. Past works~\cite{johnson2016perceptual}\cite{wang2018esrgan} declare the distance calculated at the high-level embedding space is more similar to human perception since the pre-trained  \( \Psi \) can omit the information that is helpless to human recognition. The introduced perceptual loss works as minimizing pixel loss of some compacter images without trivial information. Here we formulate the perceptual loss in terms of widely used MSE loss in the feature manifold as

\begin{equation}
    \min_{\theta_F} || \Psi(F(X; \theta)) - \Psi(Y) ||^2,
\end{equation}

where \( F\) is the generator with parameters \( \theta_F \) that learns to transform the input images \( X \) into target image \( Y \), and the weights of \( \Psi \) are fixed during the training phase.
However, the generated images that using the original pre-trained networks usually  have various artifacts. Here we assume the representation extracted from \( \Psi \) is not powerful enough to represent images due to the distribution divergence on pre-trained classification dataset and transformation dataset.
One straightforward modification is to fine-tune
the pre-trained \( \Psi \) for image transformation tasks, but it tends to be unpractical due to the lacking labeled dataset.

Here we introduce the online contrastive learning to perform simultaneous learning on both the pre-trained \( \Psi \) and generator \( F \). 
It aims to learn the distinctiveness where the similar objects are similar in feature space and different objects are different in feature space.
By learning in the self-supervised manner, it does not need the categorical label during the training.
Inspired from the unsupervised learning methods~\cite{chen2020simple} in the recognition tasks, the triplet is constructed using the random crop function \( f_c \) on instance images, which produces different cropped results at each call. The final loss function can be formulated as:
\begin{equation}
\min_{\theta_\Psi} \max(||\Psi(f_{c}(Y)) - \Psi(f_{c}(Y))||^2_2 - ||\Psi(f_{c}(Y)) - \Psi(f_{c}(\tilde{X}))||^2_2 + \mbox{margin}, 0),
\end{equation}

where \(\theta_\Psi\) is the parameters of pre-trained networks and we set \(1\) as the margin between the positive pairs and negative pairs.
During the training, the optimization of \(F\) and optimization of \(\Psi\) are conducted simultaneously and two different optimizers are used.
However, it is difficult to find a meaningful triplet from randomly sampled images, thus we apply the gradient accumulation when updating the parameters of \(\Psi\).
To be specific, we update the parameters of \(\Psi\) after \(K\) iterations that \(F\) has forwarded.
A more general understanding of this process can be to learn a pre-trained network to distinguish the generated images from the ground truth.
Thus the pre-trained networks \(\Psi\) can be thought as a discriminator used in the RealisticGAN~\cite{jolicoeur2018relativistic} except the weight of discriminator is transferred and the training difficulties at the initial stage are greatly reduced.

\subsection{Feature Selection as Fine-Tune}
In the previous section, we introduce the online contrastive learning that updates the parameters of the pre-trained networks \(\Psi\) during training.
However, it should be noted that such operation will become overfitting easily and the representation will be classification related only.
Hence the contrasting learning, as well as the generator optimizing will become difficult to convergence due to the adversarial learning  degenerated. 

To overcome this, we introduce a non-linear feature selection layer \(\Phi\) after the pre-trained network \(\Psi\) and freeze the parameters of \(\Psi\) at the fine-tuning process.
The feature selection layer consists of two convolution layers with \(1 \times 1\) kernel size, and one activation layer is inserted between them.
Such architecture is also used in the other representation networks, e.g., SimCLR~\cite{chen2020simple}.
The difference is that in our work, the parameters of the pre-trained network are frozen, and only the parameters of the feature selection layer are trainable.
With the introduced feature selection layer, we can combine the features from different channels and learn to activate the features per-channel using the feature selection layer.
Thus the features related to the images difference are activated and the representation is further disentangled with less disturbance from irrelevant features.  
Besides, a channel reduction is also used to further compress the dimension of output features.

\subsection{Task-Oriented Disentanglement}
Different from disentangling irrelevant dimensions from extracted representation, we further extend it into decomposing perceptual relevant factors, e.g., color, sharp, and other perceptual factors that affect images, called task-oriented disentanglement.
To implement this, we construct instance triplet samples in the online contrastive learning, but the negative samples is generated from \(Y\) using the task-specific distortion.
For example, the perceptual factor: colorful accuracy is hard to be measured explicitly, but a human can easily distinguish which image is more accurate in color between two distorted images when given a reference image, even if the given reference image is blurred or affected by other factors.
That is to say, human disentangle perceptual factors using contrastive samples.
Inspired by this intuition, we introduce Task-Oriented Disentanglement to separate each perceptual factors from networks implicitly.

More specifically, we disentangle the perceptual relevant factors by constructing anchor samples from the target images \( Y \) using specific distortion \( f_d \) randomly.
where the used distortion is related to the separated factor only, then online contrastive learning is performed with the representation network \( \mathbf{E} \) which is composed of online contrastive learning and feature selection as fine-tune and is optimized via minimizing
\begin{equation}
    \max(|| \mathbf{E}(f_d(f_{c}(Y))) - \mathbf{E}(f_{c}(\tilde{X}))||_2^2 - || \mathbf{E}(f_d(f_{c}(Y)) - \mathbf{E}(f_{c}(Y))||_2^2 + \mbox{margin}, 0).
\end{equation}
The whole convergence process is also illustrated in Figure ~\ref{fig:tradeoff}.
It can be concluded as firstly maximizing the distance between \( \mathbf{E}(f_d(f_{c}(Y))) \) and \( \mathbf{E}(f_{c}(Y)) \) to optimize \( \mathbf{E} \) where the representations are varied due to the difference in factors, then minimizing the distance between \( \mathbf{E}(f_d(f_{c}(Y))) \) and \( \mathbf{E}(f_{c}(\tilde{X})) \) to optimize \( \mathbf{E} \), which are similar to enlarge the distance between the \( \mathbf{E}(f_{c}(\tilde{X})) \) and \( \mathbf{E}(f_d(f_{c}(Y))) \) in the specific dimensions that related to the perceptual factors.

\section{Experimental Results}
In this section, we conduct three different experiments to validate the performance of our introduced methods, and analyze the relations between the perceptual quality.
Furthermore, we explore the relationship between perceptual quality and different settings.
Theses three experiments are performed on the Season image transfer~\cite{huang2018multimodal}, RAW low-light image illumination~\cite{chen2019seeing}, and RAW image super-resolution~\cite{zhang2019zoom} respectively.
Before training, we apply the data augmentations including random flipping, random rotation, and random cropping on the above datasets.
During training, we use Adam optimizer to update the networks with parameters ($\beta_1=0.9, \beta_2=0.999$).
The learning rate 1e-4 is used during the whole training and the mini-batch size is 1.
Moreover, all experiments are conducted using PyTorch 1.4 and CUDA 10.0 on Ubuntu 18.04 with 8x2080TI (11GB version).
The source code of our implementation is available at supplementary material.

\subsection{Season Image Transfer}
\begin{figure}[htbp]
\centering
\vspace{-10px}
\includegraphics[scale=0.17]{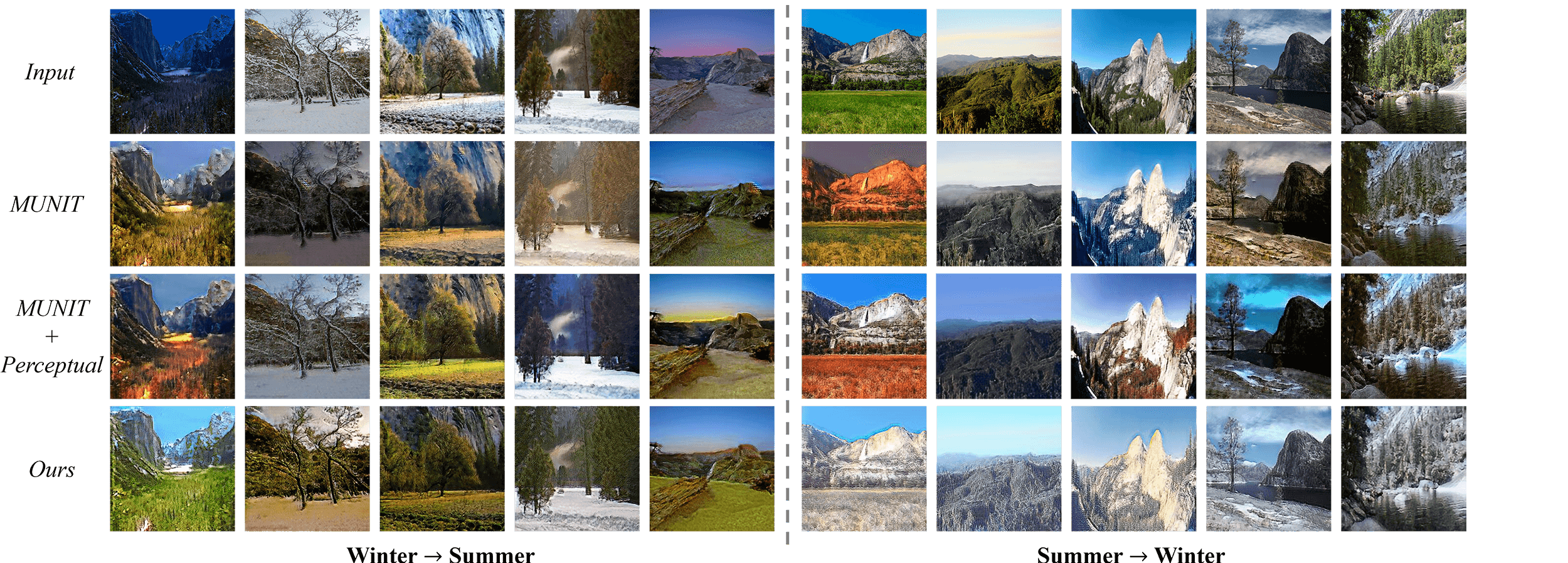}
\vspace{-12px}
\caption{Transformation results comparisons in Winter $\rightarrow$ Summer and Summer $\rightarrow$ Winter}
\vspace{-8px}
\label{fig:wtos}
\label{fig:wintersummer}
\end{figure}

Season image transfer is one of the most representative tasks in the unpaired image transformations.
During learning, the transformation network is trained on a set of winter images and a set of summer images without corresponding relations, and GAN based networks are usually used to learn the cycle relation, i.e., winter \(\rightarrow\) summer \(\rightarrow\) winter.
In order to enhance the quality of generated images, perceptual learning is applied to maximize preserve the perceptual similarity between the original images \(X\) and transformed images \(\tilde{X}\).

Here we utilize the MUNIT proposed by Huang et al.~\cite{huang2018multimodal} as the baseline of our methods.
More specificly, for online contrastive learning, we construct instance triplet samples between the source images and generated images as \(<f_{c}(X), f_{c}(X), f_{c}(\tilde{X})>\), where \(f_{c}\) is the random crop operations and the triplet is used as \(<\mbox{Anchor}, \mbox{Positive}, \mbox{Negative}>\).
With such settings, the online contrastive learning aims to maximize the distance in a feature space that only related to the season and minimize the distance in a feature space that unrelated to seasons.
In other words, the representation network is optimized to focus on more scenes that most affect the human recognized perception in winter and summer.
Since there existed recognizable divergence between the winter images and summer images, we only update the perceptual network after every 100 iterations of the generator, and we do not use the feature selection and task-oriented augmentation.
The generated results are shown in Figure~\ref{fig:wintersummer}.
In the left part of Figure~\ref{fig:wintersummer}, we show the results of Winter \(\rightarrow\) Summer, and it is clear to see that applying the perceptual can generate brighter results but have no distinct effects in scene changing.
However, after applying the perceptual with our introduced online contrastive learning, the generated results look like containing more green plants with fewer white snow in the scenes.
The results differences also exist in the right part of Figure~\ref{fig:wintersummer} which shows the Summer \(\rightarrow\) Winter results.
Therefore, we can conclude the proposed online contrastive learning is beneficial for preserving the perceptual similarity between images even when they belong to different seasons.

\subsection{RAW Low-light Image Enhancement}
\begin{figure}[htbp]
\centering
\includegraphics[scale=0.19]{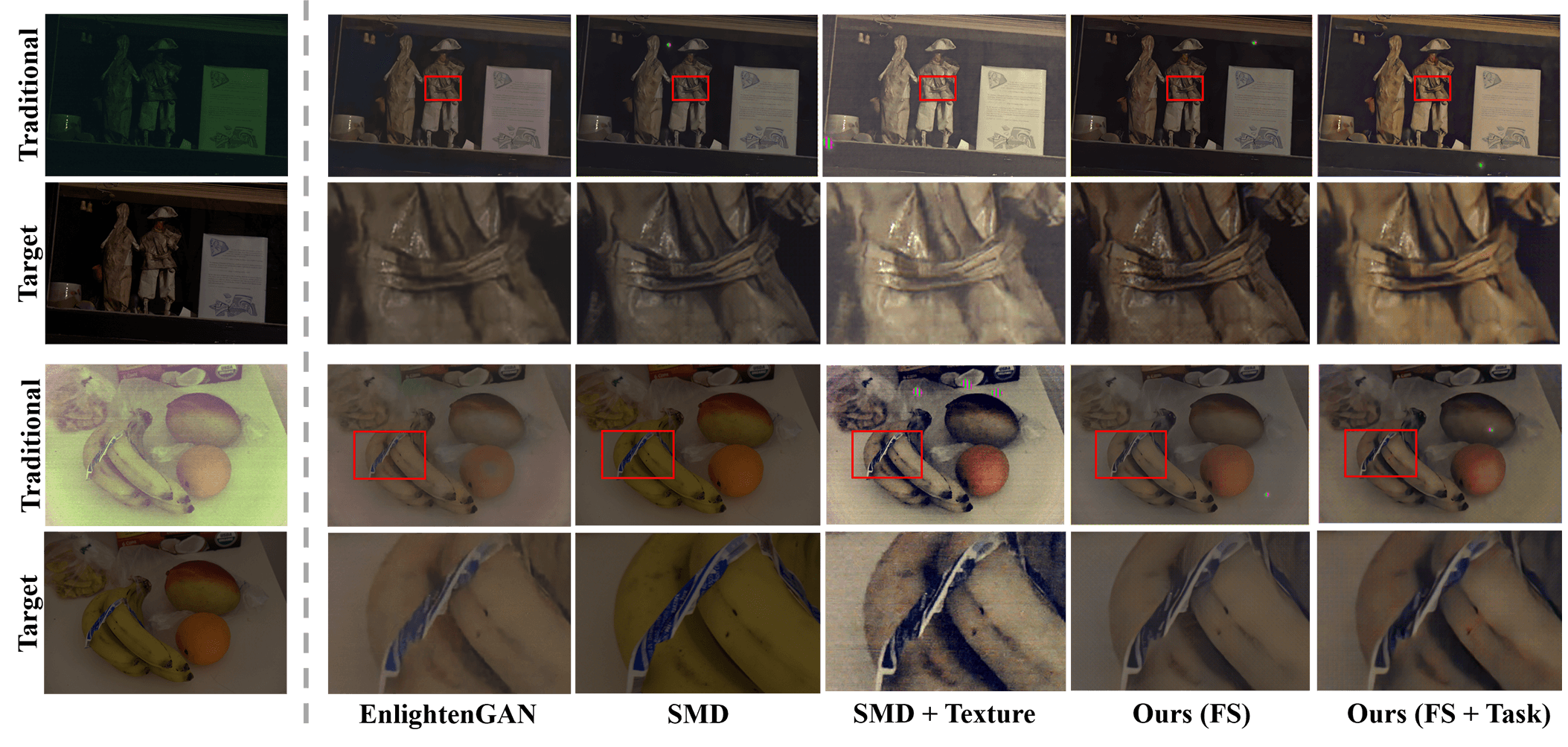}
\vspace{-14px}
\caption{Transformation results comparisons in low-light images enhancement task}
\vspace{-10px}
\label{fig:dark}
\end{figure}

In the area of low-light enhancement, RAW images have got lots of attention in recent works~\cite{chen2018learning}\cite{chen2019seeing}.
Compared with the previously used RGB images, RAW images can provide more information and have less information loss since they are generated from CMOS sensors directly.
However, these images have notable differences in visualization than normal images that humans acceptable, e.g., color space offset, optical noise, optical distortion, and so on.
Therefore, enhancing the perceptual quality of generated normal-light images have crucial practical significance.

Here we adapt the SMD proposed by Chen et al.~\cite{chen2019seeing} as the baseline of our methods, as well as the EnlightenGAN~\cite{jiang2019enlightengan} for comparison, which is the state-of-the-art methods in low-light image enhancement. SMD utilizes the extracted features from different layers of pre-trained VGG19 network,
and it calculates the distance between the illuminated images and target images in the extracted feature spaces.
The final loss of training is the combination of VGG19 loss and L1 loss.
However, as the Figure~\ref{fig:dark} shows that the generated results using such loss function tend to have uneven illumination appearance.
In order to boost the visual quality of generated results, we construct task-oriented instance triplet samples to fine-tune the pre-trained VGG19 networks with online contrastive learning.
In Table~\ref{tab:light} we show each component of our methods incrementally aims to provide a detailed analysis.
The evaluation metrics include pixel-based PSNR, perception related MS-SSIM~\cite{wang2003multiscale} and LPIPS~\cite{zhang2018unreasonable}.
For convenience, we use \textbf{FS} to denote whether the feature selection as fine-tune is used.
It is clear to see that our methods can not only achieve state-of-the-art performance in perception, but also can generate a result with the best visual quality and more realistic details, even compared to the handcrafted texture loss.

\begin{table}[h]
  \caption{Quantitative comparisons in low-light enhancement datasets.}
  \label{tab:light}
  \footnotesize
  \centering
  \begin{tabular}{llcclccc}
    \toprule
    \multicolumn{4}{c}{Methods}                   \\
    \cmidrule(r){1-4}
    Backbone & Loss & FS & Task &  PSNR \(\uparrow\) & MS-SSIM  \(\uparrow\) & LPIPS \(\downarrow\) \\
    \midrule
    Traditional & - & - & - & 17.096 & 0.8039 & 0.4185 \\
    EnlightenGAN~\cite{jiang2019enlightengan} & Adversarial & - & - & 20.556 & 0.9168 & 0.2525\\
    SMD~\cite{chen2019seeing} & VGG & - & - & 23.541 & 0.9147 & 0.1946   \\
    SMD~\cite{chen2019seeing} & VGG + Texture Loss & - & - & 22.147 & 0.8791 & 0.2218  \\
    Ours & VGG & \checkmark & - & 23.710 & \textbf{0.9210} & 0.1912  \\
    Ours & VGG & \checkmark & Blur & \textbf{24.138} & 0.9081 & \textbf{0.1874}   \\
    \bottomrule
  \end{tabular}
  \vspace{-10px}
\end{table}

\subsection{RAW Image Super-Resolution}
Recent works~\cite{zhang2019zoom}\cite{cai2019toward} have proven the remarkable differences existed between the real-world super-resolution problem and the simulated super-resolution, especially in the degeneration way of low-resolution images.
However, even training on their proposed super-resolution datasets can improve the effects especially applied the perceptual learning, 
two problems are also raised which cannot be ignored.
The first problem is the misalignment between the low-resolution images and corresponding high-resolution images during the collection.
The second problem is the color space divergence between RAW images and RGB images.
Both affect the final results seriously and make the pixel-based loss function and contextually related loss function working defectively.
In the left part of Figure~\ref{fig:tradeoff} we show the current trade-off solution that adjusts weights between L1 loss and contextual Loss to get the balance between color and sharpness.

\begin{figure}[htbp] 
\centering
\includegraphics[scale=0.19]{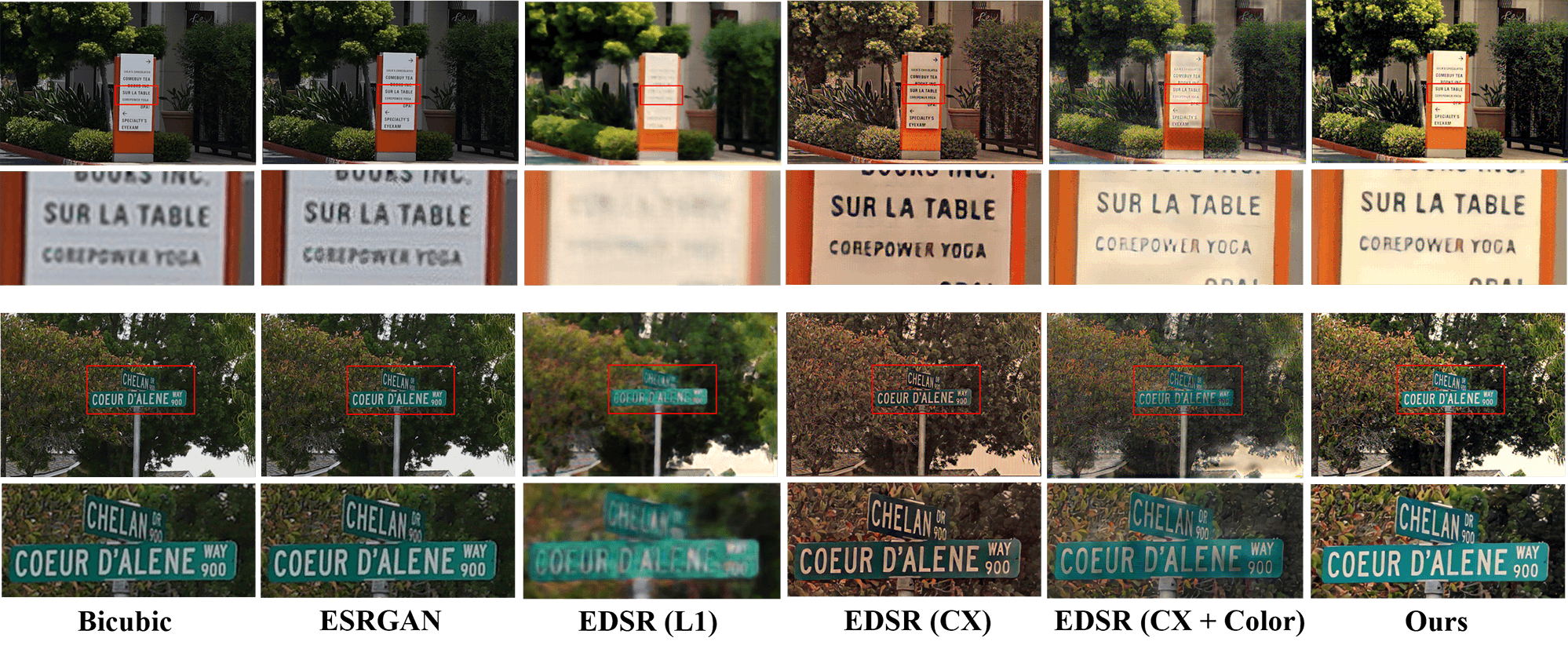}
\vspace{-10px}
\caption{Transformation results comparisons in RAW image super-resolution.}
\label{fig:srraw}
\end{figure}

To address the weight adjustment problem that existed for long, we apply the online contrastive learning to the representation network that contextual use.
It should note that even the contextual loss calculates the distance in the feature contextual spaces instead of euclidean spaces, our proposed online contrastive learning is also robustness enough since the triplet loss is relative to the difference between \(<\mbox{Anchor}, \mbox{Positive>}\) and \(<\mbox{Anchor}, \mbox{Negative}>\) instead of the difference between samples.
In table~\ref{tab:super} we show the quantitative performance comparisons between the baseline methods and our methods.
Here we apply the random color jitter to the high-resolution images as the negative samples, which makes the results with contextual only to be more colorful.
The qualitative visual results is also shown in Figure~\ref{fig:srraw}.
It is easy to conclude that our method can get the best trade-off between the sharpness and color, even no elaborate weights are used.
Noted that since the images used for validation have obviously misalignment so that the pixel-based metrics like PSNR have limited reference values.

\begin{table}
  \caption{Quantitative performance compassion between different super-resolution methods and ours.}
  \label{tab:super}
  \centering
  \footnotesize
  \begin{tabular}{llccllccc}
    \toprule
    \multicolumn{4}{c}{Methods}                   \\
    \cmidrule(r){1-4}
    Backbone & Loss & FS & Task &  PSNR \(\uparrow\) & MS-SSIM \(\uparrow\) & LPIPS \(\downarrow\) \\
    \midrule
    RGB + Bi-cubic & - & - & - & \textbf{18.556} & \textbf{0.6767} & 0.5718 \\
    RGB + ESRGAN~\cite{wang2018esrgan} & Adversarial & - & - & 18.498 & 0.6756 & 0.4389 \\
    EDSR~\cite{lim2017enhanced} & L1 & - & - & 15.705 & 0.5612 & 0.7146 \\
    EDSR~\cite{lim2017enhanced} & Contextual & - & - & 13.517 & 0.5740 & 0.5068 \\
    EDSR~\cite{lim2017enhanced} & Contextual + Color & - & - & 15.396 & 0.4950 & 0.4687 \\
    Ours & Contextual & \checkmark & Color & 14.455 & 0.5870 & \textbf{0.3913}  \\
    \bottomrule
  \end{tabular}
  \vspace{-10px}
\end{table}

\vspace{-10px}

\section{Conclusion}
In this paper, we argue that even though the perceptual learning approaches using perceptual losses on pre-trained features can capture perceptual information better than the approaches used pixel-wise losses, they also bring irrelevant information into the image transformation tasks.
We then introduce an online contrastive learning scheme to fine-tune the pre-training so that the learned representation can better represent the relationships between the results and the target images.
Specifically, we propose a feature selection layer while freezing the pre-training to preserve the natural image statistics from pre-training and reduce the irrelevant features.
Furthermore, we construct task-oriented triplet samples during the fine-tuning, which drive the feature selection layer to be more sensitive to the task related statistics.
Finally, the proposed disentangled representation can achieve more realistic results in many image transformation tasks.
Our further work will focus on disentangling the representation of human perception with finer controlling during image transformation.

\medskip
\clearpage

\appendix

\section{Additional Network Details}
\begin{figure}[htbp]
\centering
\includegraphics[scale=0.16]{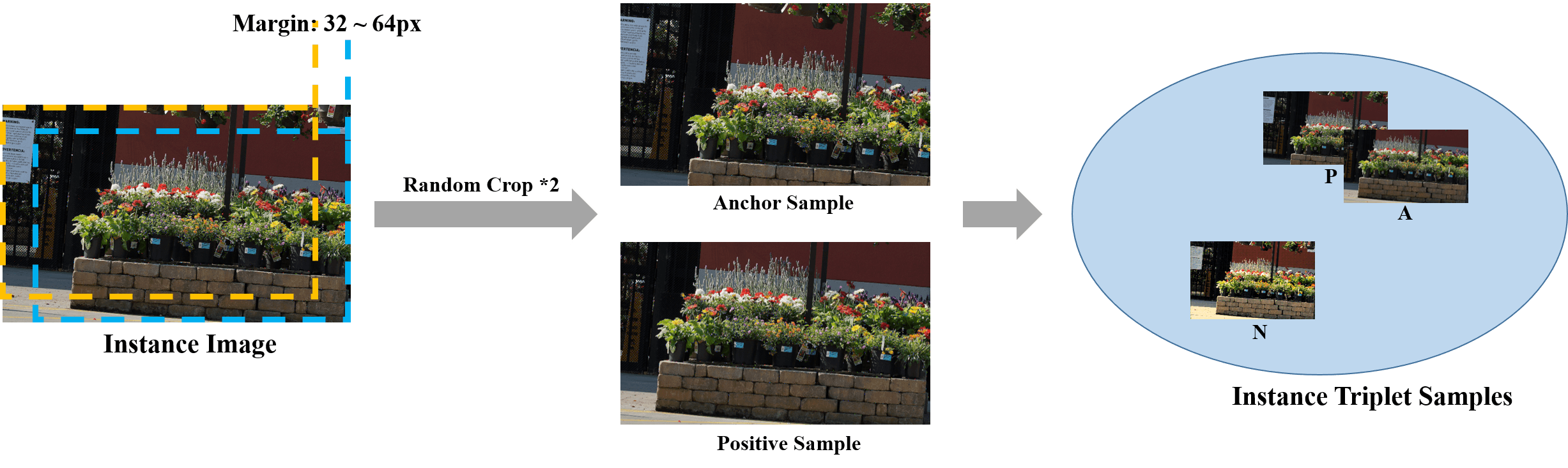}
\caption{Visualization of constructing instance triplet during online contrastive learning.}
\label{fig:triplet}
\end{figure}

\paragraph{Instance Triplet.} Construct instance triplets takes an essential role in online contrastive learning.
It utilizes the self-similarity properties to enhance the learning that distinguishes the positive samples from the anchor samples.
To help readers better understand it, We illustrate its process in Figure~\ref{fig:triplet} with the random crop function in two times.

\begin{figure}[htbp]
\centering
\includegraphics[scale=0.18]{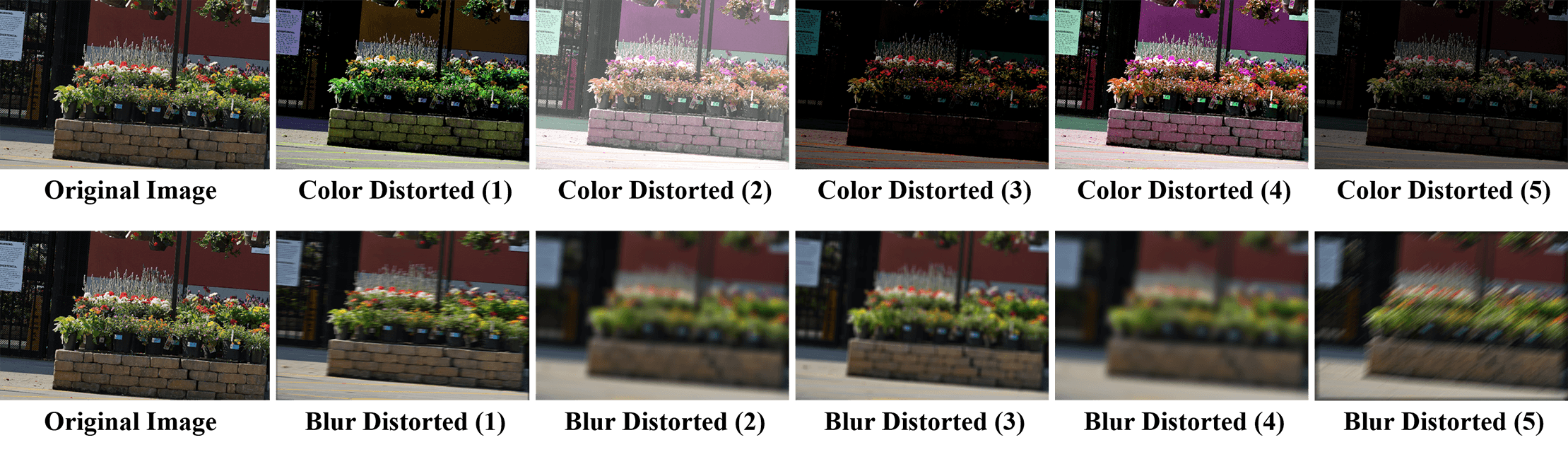}
\caption{Visualization of constructing task-oriented instance triplet during online contrastive learning.}
\label{fig:distorted}
\end{figure}

\paragraph{Task-Oriented Instance Triplet.}
Different from the online contrastive learning that depends on the self-similarity, the task-oriented disentanglement focus more on the specific perceptual factors instead of differences between the target images and generated images.
To achieve finer disentanglement during the contrastive learning, we apply the different distortion algorithms in the target images as the anchor samples.
To help readers better understand it, we illustrate the distorted results in Figure~\ref{fig:distorted} with two different distortion algorithms.

\clearpage

\section{Compared Handcrafted Losses Details}
In order to demonstrate our method with the handcrafted loss functions that can disentangle specific perceptual factors, 
we have conducted several comparison in the experiments.
In this section we will describe the details of the used handcrafted loss functions: color loss and texture loss proposed by Ignatov et al.~\cite{ignatov2017dslr}.

\paragraph{Color Loss.}
We use the color loss function to measure the difference of images in the brightness, contrast and color instead of pixel itself.
Denote the transformed image and the target image as  \(\tilde{X}\) and \(Y\), they are first processed by the  Gaussian Blur as \(G(\cdot)\) for removing high-level details, e.g., edges and textures.
With such operation, the left parts \(G(\tilde{X})\) and \(G(Y)\) only contain the information related to brightness, contrast, and color.
Hence the distance in the color and other related perceptual factors could be calculated via the Euclidean distance as:
\[
\mathcal{L}_{{color}}=||G(\tilde{X}) - G(Y)||^2_2.
\]

\paragraph{Texture loss.}
Similar to the motivation of color loss, texture loss are calculated in the grayscale version of two images, which aims to eliminate the effect of color.
The formula for texture loss is:
\[
\mathcal{L}_{{color}}=||\tilde{X}_{gray} - Y_{gray}||^2_2.
\]

\newpage
\section{Additional Result Images}
\begin{figure}[htbp]
\centering
\includegraphics[scale=0.3]{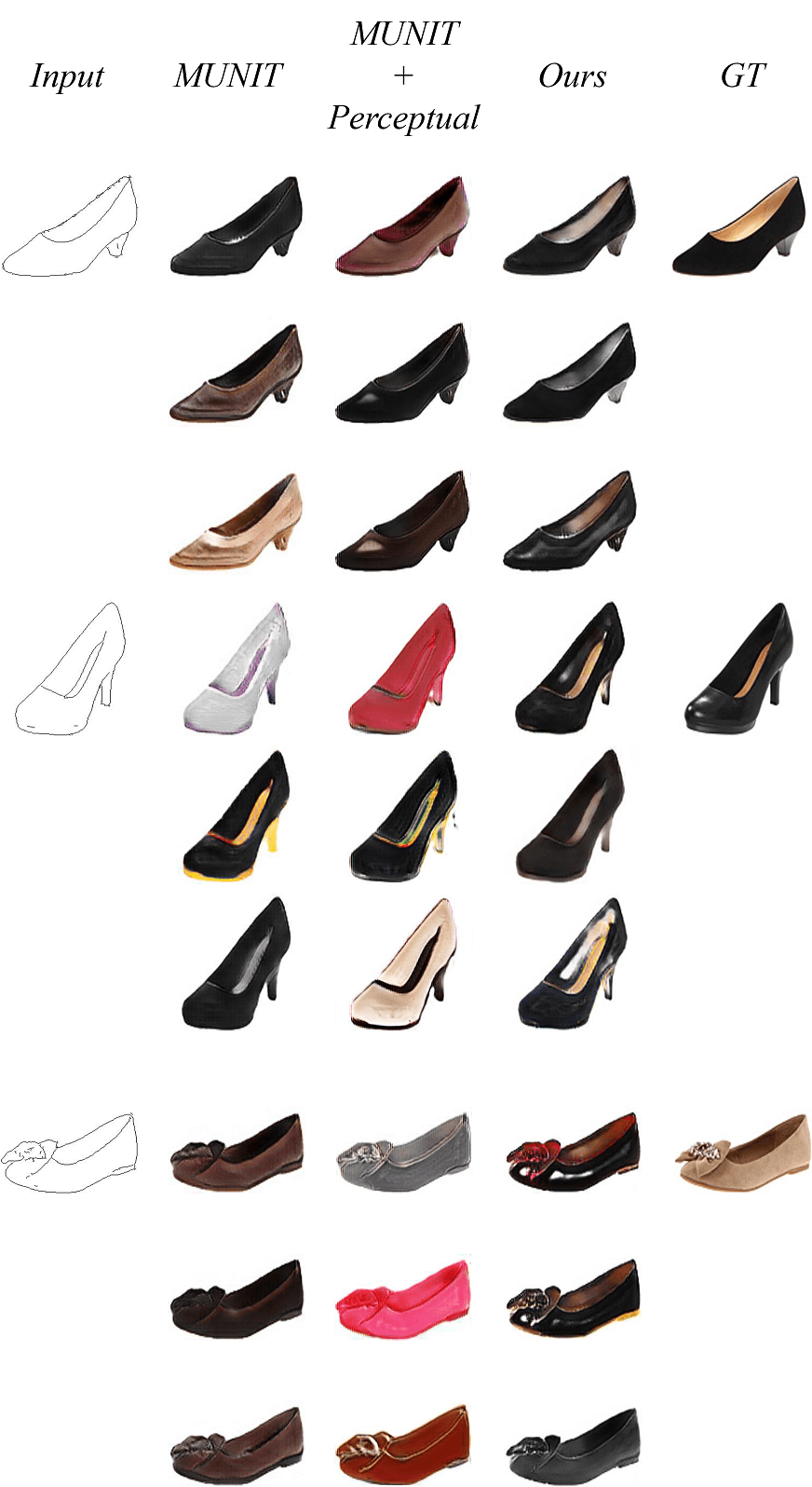}
\caption{Transformation results comparisons in edges2shoes dataset}
\label{fig:sr2}
\end{figure}

On the basis of the Season Image Transfer, we also conduct experiments in edges2shoes dataset. In edges2shoes dataset, we still utilize the MUNIT proposed by Huang et al.\cite{huang2018multimodal} as the baseline. And the method of constructing instance triplet samples is the same as the  task of Season Image Transfer. The result is shown in Figure \ref{fig:sr2}. The left is the input, and the right is the ground truth output. Each following column shows 3 random outputs from a method. And as can be seen from Figure \ref{fig:sr2}, both the \(MUNIT\) and \(MUNIT+Perceptual\) methods produce some violation artifacts, but  the result of our introduced  online contrastive learning are more in line with human perception, and have more details.

\newpage
\begin{figure}[htbp]
\centering
\includegraphics[scale=0.18]{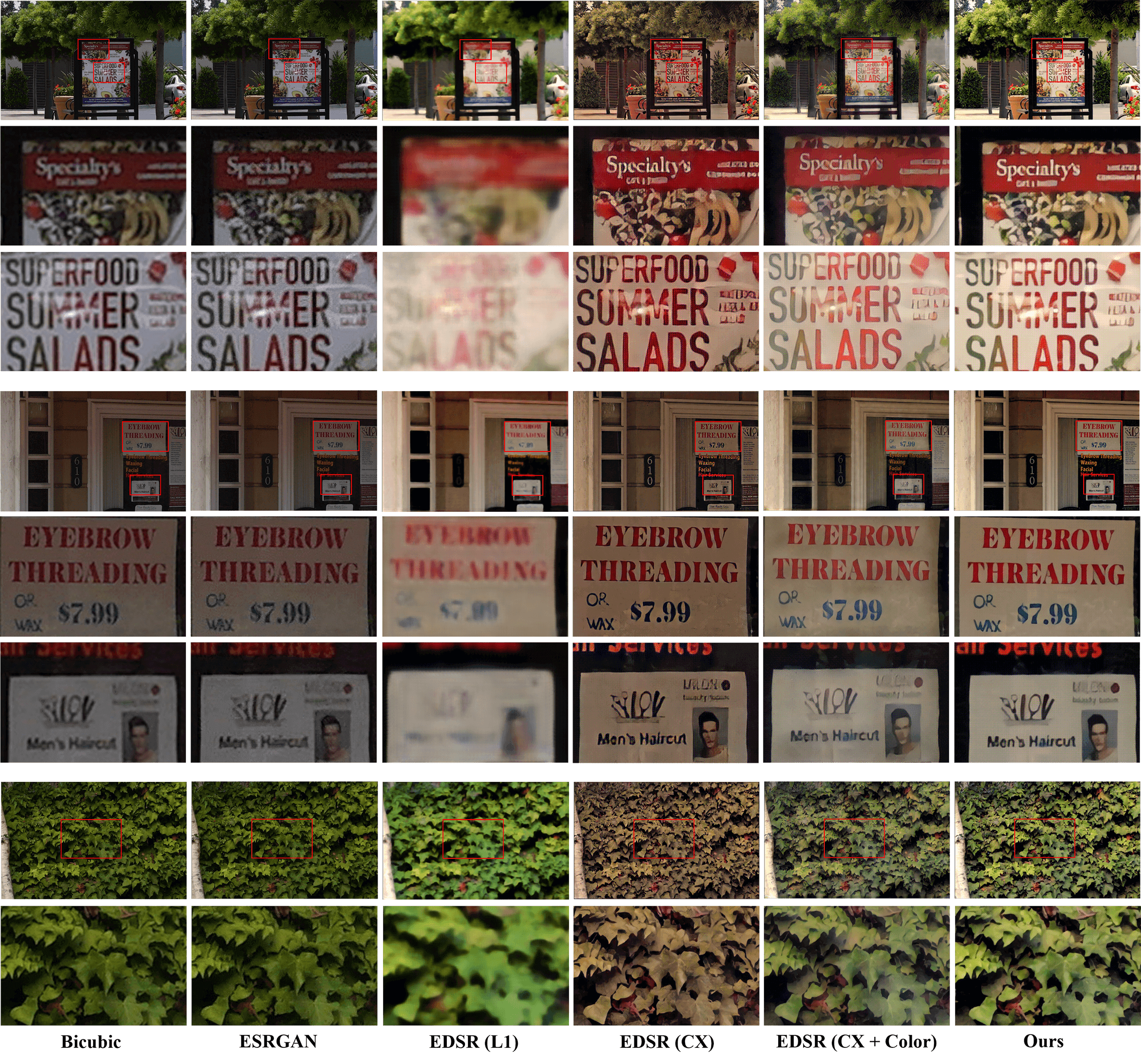}
\caption{Transformation results comparisons in RAW image super-resolution}
\label{fig:sr1}
\end{figure}

\clearpage

\section*{Broader Impact}
This paper introduces an online contrastive learning scheme to disentangle the classification-oriented pre-trained image representations for better perceptual learning in the image transformation tasks.
This framework enables the transformed images to be more realistic with fewer artifacts.
Artists, photographers, creative workers, as well as every end-user, can all benefit from it. 
It is possible this technique can be used to make more realistic ``fake'' images.
However, we believe ultimately this technique will help people to understand the mechanism of image transformation even deeper. Besides, no consequences of failure of the system exist, and no biases in the data are leveraged.

{\small
\bibliographystyle{plain}
\bibliography{neurips_2020}
}

\end{document}